\documentclass[10pt,twocolumn,letterpaper]{article}

\usepackage{wacv}              

\usepackage[accsupp]{axessibility}

\usepackage{graphicx}
\usepackage{amsmath}
\usepackage{amssymb}
\usepackage{booktabs}
\usepackage{multirow}
\usepackage{enumitem}
\usepackage[pagebackref,breaklinks,colorlinks]{hyperref}

\usepackage[capitalize]{cleveref}
\crefname{section}{Sec.}{Secs.}
\Crefname{section}{Section}{Sections}
\Crefname{table}{Table}{Tables}
\crefname{table}{Tab.}{Tabs.}

\def\wacvPaperID{1081} 
\def\confName{WACV}
\def\confYear{2025}

\begin{document}

\title{HOPE: A Memory-Based and Composition-Aware Framework for Zero-Shot Learning with Hopfield Network and Soft Mixture of Experts}

\author{Do Huu Dat$^*$\\
VinUniversity\\
{\tt\small 22dat.dh@vinuni.edu.vn}
\and
Po-Yuan Mao$^*$\\
Academia Sinica\\
\and
Tien Hoang Nguyen\\
VNU-UET\\
\and
Wray Buntine\\
VinUniversity\\
\and
Mohammed Bennamoun\\
University of Western Australia\\
}
\maketitle
\def\thefootnote{*}\footnotetext{These authors contributed equally to this work}\def\thefootnote{\arabic{footnote}}
Compositional Zero-Shot Learning (CZSL) has emerged as an essential paradigm in machine learning, aiming to overcome the constraints of traditional zero-shot learning by incorporating compositional thinking into its methodology. Conventional zero-shot learning has difficulty managing unfamiliar combinations of seen and unseen classes because it depends on pre-defined class embeddings. In contrast, Compositional Zero-Shot Learning leverages the inherent hierarchies and structural connections among classes, creating new class representations by combining attributes, components, or other semantic elements. In our paper, we propose a novel framework that for the first time combines the Modern \underline{H}opfield Network with a Mixture \underline{o}f \underline{E}x\underline{p}erts (HOPE) to classify the compositions of previously unseen objects. Specifically, the Modern Hopfield Network creates a memory that stores label prototypes and identifies relevant labels for a given input image. Subsequently, the Mixture of Expert models integrates the image with the appropriate prototype to produce the final composition classification. Our approach achieves SOTA performance on several benchmarks, including MIT-States and UT-Zappos. We also examine how each component contributes to improved generalization.  
\section{Introduction}
\label{sec:intro}

In recent years, machine learning has made substantial advancements, with innovations covering a wide range of applications such as image categorization and natural language interpretation. Among the evolving paradigms, Zero-shot Learning (ZSL) has gained attention, aiming at overcoming the challenge of model generalization to unseen categories with scarce labeled instances. Conventional ZSL methodologies use auxiliary information like class attributes and class embeddings but struggle with unseen class combinations. In this context, Compositional Zero-Shot Learning (CZSL) is emerging as a viable solution, especially in structured or high multiclass situations.
\vspace{-.5cm}
\begin{figure}[!ht]
    \centering    \includegraphics[width=0.9\columnwidth,keepaspectratio]{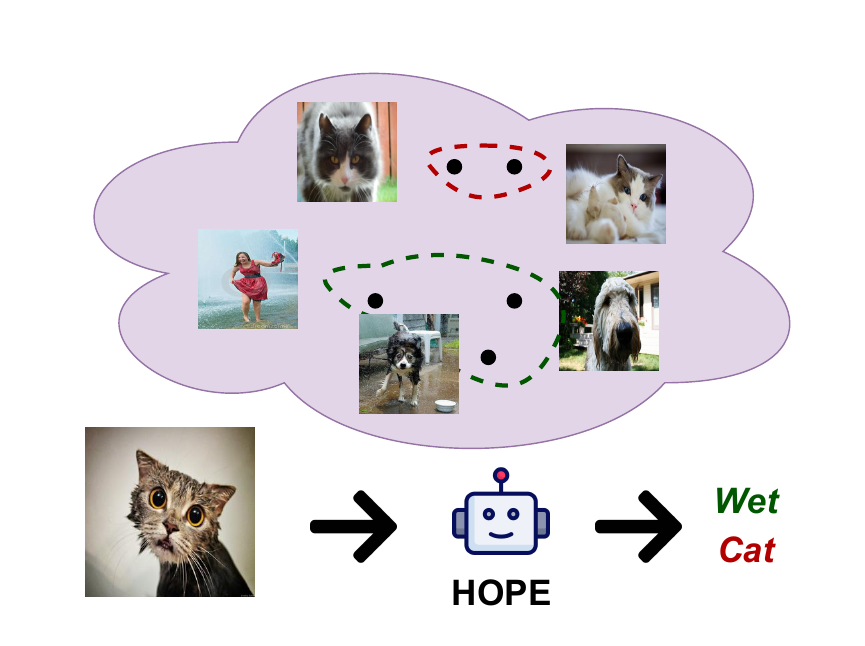}
    \caption{Given a new unseen input composition (wet cat),  HOPE retrieves analogous visual and linguistic concept embeddings from its learnable memory (wet dog, wet clothes, dry cat, cute cat, etc) and based on the additional retrieved information to make the ultimate prediction.}
    \label{overview}
    \vspace{-.5cm}
\end{figure}

Compositional Zero-Shot Learning (CZSL) is a task aimed at recognizing unseen compositional visual concepts by training models on learned concepts. Fig~\ref{overview} demonstrates the basic idea of CZSL. It primarily involves modeling attributes that interact with different objects and addressing the challenges posed by diversity, informativeness, and entanglement between visual primitives. CZSL effectively incorporates compositional reasoning, which facilitates models to accurately predict new class combinations by considering the hierarchical and structural relationships among classes. This is achieved through the generation of embeddings based on constituent components, thereby offering a robust strategy towards tackling Zero-Shot Learning (ZSL) challenges. Through this methodical approach, CZSL not only broadens the scope of recognizing new visual compositions but also significantly contributes towards enhancing the model's generalization capabilities in unseen scenarios.

A common CZSL strategy involves merging scores from different classifiers, independently trained for attributes and objects \cite{misra2017red}. Nevertheless, such separate predictions often overlook the contextual interactions originating from the combination of different primitives, termed as contextuality within compositions  \cite{misra2017red}. Previous approaches have tackled this challenge by modeling compositional label embeddings for each class, making use of external knowledge repositories like Glove \cite{mikolov2013distributed} to extract semantic vectors for attributes and objects. These vectors are subsequently concatenated through multiple layers to form the compositional label embedding. The resulting embeddings are then aligned with visual attributes in a shared embedding space, effectively transforming the recognition of unseen images into a nearest neighbor search problem  \cite{mancini2021open}. However, the extent of contextual consideration within the visual domain remains constrained. Prior investigations have also underscored the significance of discriminability in visual features, which influences generalization performance in recognizing unseen compositions. Disentanglement emerges as a favored solution, with dedicated layers for extracting intermediate visual representations of attributes and objects \cite{ruis2021independent, saini2022disentangling, li2022siamese, zhang2022learning}.

In this paper, we propose a novel methodology intended to faithfully mimic the complex process wherein humans inherently fuse various visual elements to create compositions. Our assertion on the profound impact of memory on human compositionality is further corroborated by the way individuals employ towards novel concepts. Upon encountering unfamiliar ideas or scenarios, humans frequently reason and formulate logical conjectures rooted in their pre-existing knowledge and memory \cite{crossley2023circuit}. Additionally, the interactions between diverse attributes and objects are unique, necessitating varied composition techniques. 

Motivated by this, we propose HOPE, a novel and efficient model centered on learning primitive representations, retrieving, and combining them to create compositional and joint representations.  To learn the joint representations, HOPE exploits the soft prompt from DFSP \cite{lu2023decomposed}. Furthermore, we develop a set of learnable features, initially represented by the mean embeddings of random images corresponding to the same classes, and use the Modern Hopfield Network \cite{ramsauer2020hopfield} to retrieve representations of akin attributes or objects predicated on the input image. The Modern Hopfield Network, drawing from biological concepts, aids in establishing an associative memory framework crucial for the retrieval process, emulating a human-like approach toward recognizing and categorizing new or unseen objects or scenarios. Associative memory allows the network to recall patterns or representations based on partial or noisy inputs, providing a robust mechanism for retrieval even in the face of incomplete or imperfect cues. This property is essential for real-world applications where inputs are often ambiguous or occluded. Lastly, we deploy a Soft Mixture of Experts (Soft MoE) \cite{puigcerver2023sparse} to adapt the input image embeddings based on the retrieved features. The ``soft" in Soft Mixture of Experts refers to the manner in which the model allocates the handling of inputs among different ``experts" within the model. Unlike a ``hard" Mixture of Experts model where each input is directed to exactly one expert based on the gating network, a Soft Mixture of Experts model allows for a more nuanced allocation, permitting each input to be handled by multiple experts with varying degrees of weighting determined by the gating network. This soft assignment enables a more flexible and nuanced handling of inputs, which could be particularly beneficial in complex or multifaceted tasks, providing a form of regularization and a richer representation of the data as it gathers insights from multiple experts for each input.  The sparsity of the Soft MoE model is perceived as a form of specialization, with certain experts dedicated to managing specific attributes. 

To summarize, the primary contributions of this paper are as follows: 
\begin{itemize}
    \item We introduce an innovative framework - HOPE, inspired by human memory, leveraging external references to effectively recognize novel combinations using exclusive primitives.
    \item For the first time, our proposed framework employs a Modern Hopfield Network and Mixture of Expert models for Compositional Zero-Shot Learning (CZSL), capitalizing on their synergy to enhance the classification of unseen compositions.
    \item We also devise additional loss functions aimed at progressively augmenting the memory’s composability while preserving the primitive clustering. 
\end{itemize}
\noindent
To evaluate these contributions, we conduct extensive experiments examining the influence of memory design and auxiliary loss functions on the overall model performance, alongside the effectiveness of other elements within our methodology.
\section{Related Works} 

\subsection{Compositional Zero-Shot Learning}
\hspace{1pc}Compositional Zero-Shot Learning (CZSL) \cite{li2020symmetry, mikolov2013distributed, misra2017red, nagarajan2018attributes} is a task that mimics the human ability to imagine and identify new concepts based on prior knowledge, marking it as a crucial part of Zero-Shot Learning (ZSL) \cite{chen2023duet, guo2020novel, guo2023graph, lampert2013attribute, li2021generalized, liu2021goal,liu2022towards}. While traditional ZSL uses attributes to recognize unseen objects, CZSL views classes as mixtures of states and objects.

Initially, CZSL methods involved training a classifier for identification along with a transformation module to change states or objects \cite{misra2017red,nagarajan2018attributes}. However, newer strategies use two distinct classifiers to identify states and objects individually \cite{karthik2022kg,li2022siamese,li2020symmetry,misra2017red}. Some methods even combine the encoded attribute/state and object features using late fusion with a multi-layer perceptron \cite{purushwalkam2019task}. Li et al. \cite{li2022siamese} brought contrastive learning into CZSL, designing a siamese network to identify states and objects in a contrastive space. Other strategies \cite{nagarajan2018attributes, nan2019recognizing} aim at representing compositions together, learning an embedding space to map compositions like in ZSL. Moreover, methods have been using graph networks to represent the relationship between states and objects and to learn their compositions \cite{mancini2022learning, ruis2021independent}. Lately, the attention mechanism has been modified to disentangle compositions into primitives and compute new compositions \cite{kim2023hierarchical, li2023distilled, hao2023learning}. Nihal et al. \cite{nayak2022learning} pioneered the use of a VLM (Vision-Language Model) for CZSL, substituting prompt classes with a learnable combined state and object vector representation.

Additionally, the challenge of handling compositions during testing is not just restricted to the test data compositions, but often extends to real-world scenarios, accounting for all potential compositions. Contrary to the earlier closed-world setting (where the system only encounters scenarios it was trained on), some research aims to tackle the open-world situation (where the system may encounter completely new, unseen scenarios) by using external knowledge to eliminate unlikely compositions \cite{karthik2022kg, mancini2022learning, mancini2021open, wang2023learning}.

\subsection{Associative Memory}
In our proposed approach, using Associative Memory is key for efficient storage and retrieval of learned patterns, enabling precise mapping of new compositional representations. This capability allows our framework to efficiently navigate the complexities of open-world scenarios. This enables it to deduce the characteristics of unseen compositions, significantly enhancing the accuracy and robustness of CZSL in real-world applications.

Looking more closely into the technical aspects, associative memory networks play a central role in linking an input with the most closely resembling pattern, focusing on both the storage and retrieval of patterns. 
The classic Hopfield Network \cite{hopfield1982neural} stores multidimensional vectors as memories using recurrent dynamical systems. This approach facilitates data clustering by creating fixed-point attractor states.
Yet, a significant limitation lies in its restricted memory capacity, which can only hold around 0.14 times the dimensionality ($d$) in random memories for a d-dimensional data domain. This poses a challenge for clustering, especially when the number of clusters should not be directly related to data dimensionality. 
To address the drawbacks associated with the classical Hopfield Network, Krotov and Hopfield introduced an updated version known as the Modern Hopfield Network, or Dense Associative Memory (Dense AM)\cite{krotov2016dense}.This improved model incorporates rapidly advancing non-linearities (activation functions) into the dynamical system, which results in a denser memory arrangement and significantly boosts the memory capacity. This smart design especially benefits in high-dimensional data spaces. Beside, some of the activation functions employed in Dense AMs have the potential to lead to power-law or even exponential memory capacity \cite{demircigil2017model, lucibello2023exponential}.

Moreover, as pointed out by Ramsauer et al., the attention mechanism inherent in transformers\cite{vaswani2017attention} can be interpreted as a particular instance of Dense AMs when utilizing the softmax activation function \cite{ramsauer2020hopfield}. They illustrated the capacity to store an exponentially large number of patterns, retrieve patterns with a single update, maintain an exponentially small retrieval loss, and even can learn the memory.

\subsection{Sparse Mixture of Experts}
Sparsely-gated Mixture of Experts (MoE) expands the capabilities of associative memory, leading the further improvement in managing and interpreting complex data structures. This is especially notable in the context of CZSL.
The MoE model has been groundbreaking, displaying substantial improvements in model capacity, training time, or model quality through the incorporation of gating. Following this, the Switch Transformer \cite{fedus2022switch} simplified the gating process by choosing only the top expert per token using a softmax over the hidden state, which displayed better scaling compared to earlier efforts. However, a common necessity across these improvements has been the use of an auxiliary loss to actively promote balancing within the model. This auxiliary loss needs meticulous weighting to prevent overshadowing the primary loss, yet it doesn't ensure a perfect balance, necessitating a hard capacity factor. This scenario might lead to many tokens remaining unprocessed by the MoE layer. The introduction of Hard MoE \cite{gross2017hard}with a singular decoding layer showed efficient training yielding positive outcomes on large-scale hashtag prediction tasks. 

Furthermore, Base Layers \cite{lewis2021base} devised a linear assignment strategy to maximize token-expert affinities while ensuring an equitable distribution of tokens to each expert. Recently, several innovative methods have surfaced, proposing ways to selectively trigger token paths across the network in various domains including language \cite{lepikhin2020gshard, fedus2022switch}, vision \cite{riquelme2021scaling}, and multimodal models \cite{mustafa2022multimodal}. This diverse array of strategies underscores the potential of leveraging sophisticated models like MoE in addressing the complex challenges posed by the open-world setting of CZSL, making a compelling case for its inclusion in our proposed framework. 

\begin{figure*}[!ht]   \centering\includegraphics[width =0.9\textwidth ]{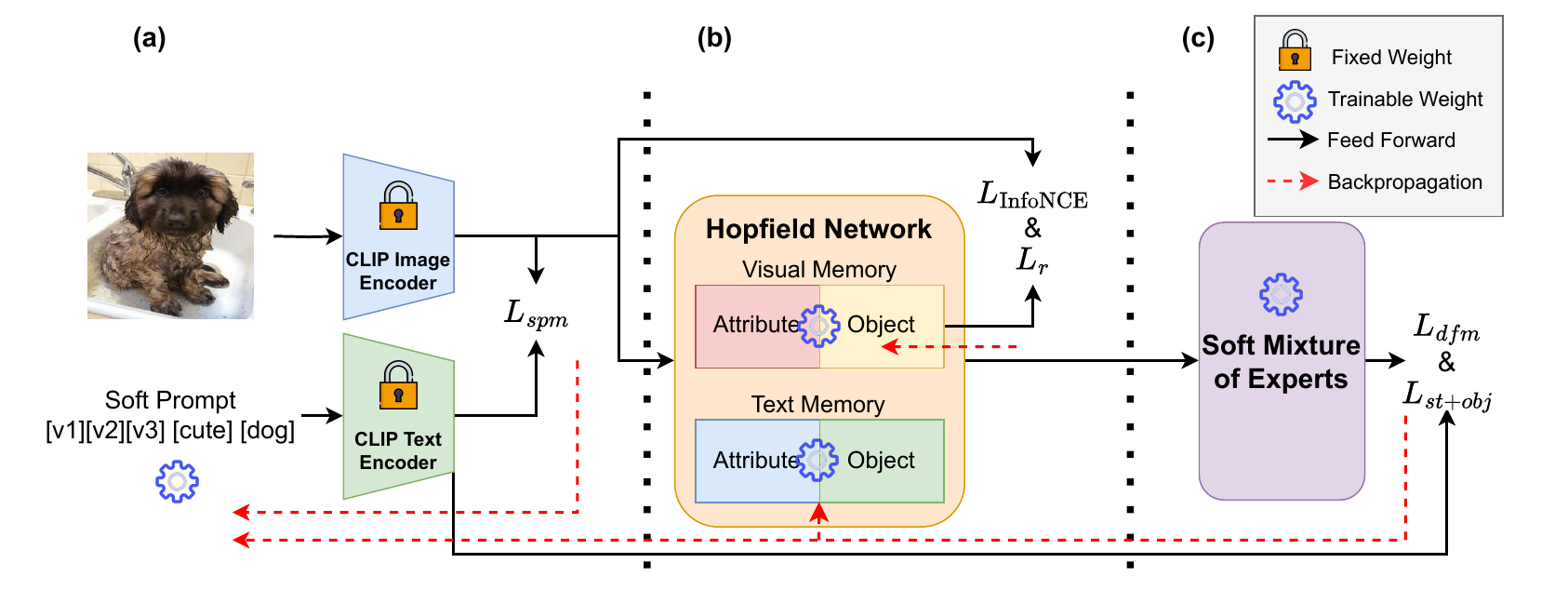}
\caption {
\textbf{Framework Illustration of HOPE.}
HOPE is a three-stage training model. (a) We train the soft prompt by calculating $L_{\text{spm}}$ between CLIP encoding. The soft prompt consists of trainable variable [V1][V2][V3] and the set of labels [cut][dog]. (b) Further, we forward the output of the image latent features to the Hopfield network to retrieve the nearest samples from each class. Then we introduce $L_{\text{InfoNCE}}$ and $L_r$ to optimize the visual memory. (c) Finally, Soft Mixture of Experts will aggregate the information and give the ultimate prediction. The prediction is trained by $L_{\text{dfm}}$ and $L_{\text{st+obj}}$.
}    
    \label{HOPE_framework}
\vspace{-.3cm}
\end{figure*}
\section{Problem Formulation}
Compositional Zero-Shot Learning is a classification problem that aims to train a model using a limited set of attribute-object combinations, enabling the recognition of images with unseen combinations. 
To elaborate, consider an attribute set $A = \{a_0, a_1, \ldots, a_n\}$ and an object set $O = \{o_0, o_1, \ldots, o_m\}$. By composing these two labels, a composition set is formed as $C = A \times O = \{(a_0, o_0), (a_1, o_0), \ldots, (a_n, o_m)\}$. The set $C$ is used to label images with seen classes $C_s\subset C$ and unseen classes $C_u\subset C $ where $C_s\cap C_u=\emptyset$ and each image only has one $c\in C$ label. This division is also used to separate training $\mathcal{D}^{tr}= \{(X_{tr},C_{tr})\}$ and testing $\mathcal{D}^{test} = \{(X_{test}, C_{test})\}$ datasets, where $X_{tr}$ is labelled with seen classes $C_s$ only, so $C_{tr}\subseteq Powerset({C_s})$.
Whereas $X_{test}$ are labeled with $C_s$ and ${C_u}$ classes. \cite{karthik2022kg} defined the open-world evaluation as $C_{test}$ labels are from $C_s \cup {C_u}$, while for closed-world evaluation $C_{test}$ labels are from $C_s \cup {C_u}'$ where ${C_u}' \subset C_u$.
Compositional Zero-Shot Learning aims to train a classification model $f_\theta(X)$ on $\mathcal{D}^{tr}$ but it should also be capable of correctly predicting $\mathcal{D}^{test}$.

\vspace{-.2cm}
\section{HOPE}
We proposed HOPE, a retrieval-support method that understands the input image via similar seen state or object embeddings and then consensus to make a new embedding. Specifically, our proposed model (HOPE) consists of three primary steps: (1) soft prompt module, (2) retrieval using the modern Hopfield network, and (3) prototype composition. The framework of HOPE is depicted in Fig~\ref{HOPE_framework}. More detailed explanations of our approach are provided in the following sections.




\subsection{Soft Prompt Module}
Following \cite{lu2023decomposed}, we train the soft prompt to construct the
joint representation of attribute and object. Consider an image denoted by $x_i$ and a collection of labels $C_i ={(a_i,o_i)}$, both of which are sampled from the training dataset $\mathcal{D}^{tr}$. We first transform labels into a soft prompt $P_\theta(C_i) =$ [$\theta_1$] [$\theta_2$] [$\theta_3$] [$a^i$] [$o^i$], with [$\theta_1$] [$\theta_2$] [$\theta_3$] trainable. We forward the soft prompt and the image to an off-shelf CLIP model. Then we minimize the cross entropy loss of the soft prompt module:
\begin{equation}
\resizebox{.35 \textwidth}{!} 
{$L_{spm}= - \frac{1}{C_s}\sum_{(x,y\in C_s)}log(p_{spm}(\frac{y=(s,o)}{x;\theta}),$}
\end{equation}
where class probability $p_{spm}(\frac{y=(s,o)}{x;\theta})$ is defined as:

\begin{equation}
\resizebox{.35 \textwidth}{!} 
{$
    p_{spm}(\frac{y=(s,o)}{x;\theta})= \frac{exp(f_v \cdot f_t)}{\sum _{(\bar{s},\bar{o}\in C_s)} exp(f_v \cdot f_t)}$}.
\end{equation}
Here $f_v$ and $f_t$ are the output from CLIP image and text encoder.

\subsection{Modern Hopfield Network}
In the context of CZSL, our objective is to retrieve key visual prototypes while maintaining a generalized visual representation as memory. To achieve this, we turn to the Modern Hopfield Network, which enables us to retrieve representations of other labels and keep updating these representations continually.

Going into the details, Hopfield Network has visual $M_v$ and text memory $M_t$. Each memory contains two small sections, which are donated as $M_{v}^{a}$, $M_{v}^{o}$, $M_{t}^{a}$, $M_{t}^{o}$  where $M_{v}^{a}, M_{v}^{o} \in \mathbb{R}^{|C_{tr}| \times D}$, $M_{t}^{a} \in \mathbb{R}^{|A| \times D}$ and $M_{t}^{o} \in \mathbb{R}^{|O| \times D}$, 
respectively representing the memory of attributes and objects with a latent dimension $D$.
Take constructing $M_t^a$ as an example, the linguistic prototype is obtained from the output of the frozen CLIP's text encoder with a prompt: "A photo of [attribute] object". Similarly, the prompt "A photo of [object]" is used to construct $M_t^o$. 
Furthermore, the visual prototype is calculated as the mean of multiple image embeddings produced by the frozen CLIP image encoder, based on images from the same class.
The rationale behind encoding two distinct sets of features is to capture the interplay of primitives with another primitive. Rather than representing just an attribute or object, we construct $M_v^a$ and $M_v^o$, ensuring they match the size with the training set classes, so that we can store noticeable variations of primitives for improved compositions. 

For a Hopfield Network to retrieve $l$ different patterns, we initially use a trainable linear layer to project $f_v$ into $l$ vectors, denoted by $Z_l = Z_\theta(f_v)$. Subsequently, the modern Hopfield network retrieves akin patterns, $V = [V_1, V_2, ..., V_l]$, from both $M_v^a$ and $M_v^o$, with each set containing $l/2$ patterns, respectively. This equation is adapted from the modern Hopfield Network \cite{ramsauer2020hopfield}.
\begin{equation}
    V_i = \left\{\begin{matrix}
    softmax(Z_{i} \cdot (M_v^a)^T)\cdot M_v^a, & i<l/2 \\ 
    softmax(Z_{i} \cdot (M_v^o)^T)\cdot M_v^o, & i\geq l/2 
    \end{matrix}\right.
\end{equation}
In this equation, $s_i = softmax(Z_{i} \cdot M_v^a)$ can be viewed as the probability of the vector correlating to the memory. To ensure the Modern Hopfield Network retrieves \([V_1, V_2, \ldots, V_{\textcolor{red}{l}} ]\) as meaningful information, we incorporate two auxiliary losses to realize this goal. Elaborating further, we follow the retrieval equation to compute the cross-entropy loss with the labels $(a_t , o_t)$ of the input image.
\begin{small}
\begin{align}
    \begin{split}  
        L_{r}(s) = &\sum_{i=1}^{|C_{tr}|}I(a_t)_i log \left[\frac{1}{l/2} \sum_{j=1}^{l/2}s_{j,i}\right]+ \\
        & \sum_{i=1}^{|C_{tr}|}I(o_t)_i log\left[\frac{1}{l/2} \sum_{j=l/2}^{l}s_{j,i} \right],
    \end{split}
\end{align}
\end{small}
with $I()$ is the one-hot-encoder.
Additionally, we compute the InfoNCE loss (as outlined in \cite{he2020momentum}) on the Modern Hopfield model’s attribute output and object output, respectively.

For a more comprehensive understanding, our initial step consists of generating all possible combinations of positive sets, denoted as $V^+$, from the set V. We select pairs of vectors $(V_i, V_j)$ such that $arg(s_i) = arg(s_j)$. For each of these positive sets, we proceed to calculate the InfoNCE loss using the subsequent equation:
\begin{equation}
\resizebox{.35 \textwidth}{!} 
{$
L_{\text{InfoNCE}} = - \sum_{i=0}^{l} \log \frac{\exp(f_v \cdot V^+ / \tau)}{\sum_{j=0}^{l} \exp(f_v \cdot V_{j} / \tau)}~,$}
\end{equation}
where $V$ is a set of the retrieved memory from each class which come from the Hopefield network, $V^+$ represents a $V$ having the same class with $f_v$ and $\tau$ is a temperature hyper-parameter as per \cite{wu2018unsupervised}. The underlying idea of this loss is to promote the clustering of similar attributes and objects in memory, given that the Modern Hopfield Network identifies the k-nearest neighbors to the input image, and effective clustering boosts the relevance of the retrieval mechanism.

\subsection{Prototypes aggregation.}
The significant variations in the relationships between states and objects, which may highly depend on the specific state and object in question, act as a strong driving force for our approach. When combining different attributes with an object, it is essential to customize the composition method accordingly. Given this incentive, we have adopted the Soft Mixture of Experts (Soft MoE)\cite{puigcerver2023sparse} as the compositional composer in our model. This choice allows us to infuse diversity into the model’s capacity to handle a broad spectrum of patterns and complexities present in the data.

The Soft MoE approach employs a differentiable routing algorithm, which relies on adjustable parameters for each slot to compute the importance scores for pairs of input tokens and slots. Following this, each slot uses these scores to perform a weighted summation of the input tokens. Each "expert" is essentially a unique Multi-Layer Perceptron (MLP) layer tasked with processing its assigned slot (in our experiments, we allocated one slot per expert). Ultimately, the initial importance scores are also used to combine the outputs from all the slots. In alignment with their work, we also incorporate Transformers \cite{vaswani2017attention} and substitute the second half of the MLP blocks with Soft MoE layers as the composer for our model.

\begin{table*}[!htb]
    \begin{center}
    \fontsize{9}{11}\selectfont
        \begin{tabular}{r|c c c c | c c c c | c c c c}
        \toprule
             \multirow{2}{*}{Method} &  \multicolumn{4}{c}{MIT-States} & \multicolumn{4}{|c}{UT-Zappos} & \multicolumn{4}{|c} {C-GQA}\\
             & S & U &  H & AUC & S & U &  H & AUC & S & U &  H & AUC \\
             \midrule
             CGE \cite{naeem2021learning} & 31.1 & 5.8 & 6.4 & 1.1 & 62.0 & 44.3 & 40.3 & 23.1 & 32.1 & 2.0 & 3.4 & 0.5 \\
             Co-CGE \cite{mancini2022learning} &  31.1 & 5.8 & 6.4 & 1.1 & 62.0 & 44.3 & 40.3 & 23.1 & 32.1 & 2.0 & 3.4 & 0.5 \\
             CLIP \cite{radford2021learning} & 30.2 & 45.9 & 26.1 & 11.1 & 15.8 & 49.2 & 15.6 & 5.0 & 7.7 & 24.8 & 8.4 & 1.3 \\
             CSP \cite{nayak2022learning} & 46.6 & 49.9 & 36.3 & 19.4 & 64.2 & 66.2 & 46.6 & 33.0 & 28.8 & 26.8 & 20.5 & 6.2  \\
             DFSP \cite{lu2023decomposed} & 46.9 & 52.0 & 37.3 & 20.6 & 66.7 & 71.7 & 47.2 & 36.0 & \textbf{38.2} & \textbf{32.0} & \textbf{27.1} & \textbf{10.5}\\
             \midrule
            \textbf{HOPE} & \textbf{50.5} & \textbf{54.6} & \textbf{39.9} & \textbf{23.3} & \textbf{68.4} & \textbf{73.9} & \textbf{49.1} & \textbf{37.5} & 35.8 & 30.8 & 24.5 & 9.1\\
             \bottomrule
        \end{tabular}
        \caption{\label{closed-world} Closed World Evaluation. Comparison to state-of-the-art models}
    \end{center}
\vspace{-.5cm}
\end{table*}
\begin{figure*}[!htb]
    \centering\includegraphics[width=\textwidth]{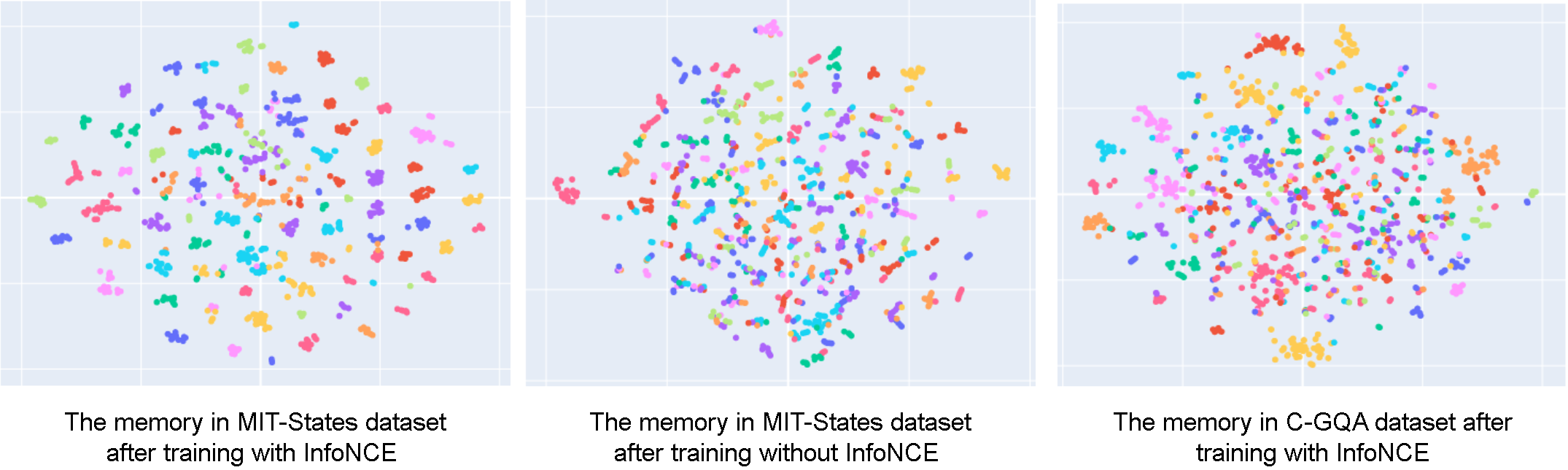}
    \caption{The visualization of attribute memory embeddings in different memory configurations and datasets}
    \label{tsne}
\vspace{-.5cm}
\end{figure*}

Given an input image embedding denoted as $f_v$ and a set of retrieved visual embeddings, namely $V_1, V_2, \ldots, V_{l}$, we also acquire the corresponding linguistic embeddings from $M_t^a$ and $M_t^o$, labeled as $T_1, T_2, \ldots, T_{l}$. For instance, when we input an image of a ``red car" into the Hopfield Network, the network produces a set of prototypes like ``red tomato," ``old car," and so forth. The textual retrieval, in this scenario, would include words like ``red," ``old," ``car," ``tomato," and others. We opt to keep these retrieved embeddings as representations of individual primitives because they are easier to combine compared to representing the entire composition, such as ``red tomato" and ``old car". Then, we concatenate the input image embeddings with the retrieved embeddings to form \([f_v, V_1, V_2, \ldots V_{l}, T_1, T_2, \ldots T_{l}]\) and feed it forward to the Soft MoE. The output features $F_v$ will be trained with $L_{\text{dfm}}$ and $L_{\text{st+obj}}$, which are defined as:
\begin{equation}
\resizebox{.35 \textwidth}{!} 
{$
L_{\text{st+obj}} = - \frac{1}{|C_s|}\sum_{(x,y)\in C_s}\log\left(\frac{\exp(F_v \cdot f_t )}{\sum_{(\bar{s},\bar{o})\in C_s}\exp(F_v \cdot f_t)}~\right),$}
\end{equation}
\begin{equation}
\resizebox{.35 \textwidth}{!} {$
    \begin{split}
    L_{dfm} =& -\frac{1}{|A|} \sum_{x,y \in C^s}log(p(\frac{y=a}{x;\theta}))\\
    &-\frac{1}{|O|} \sum_{x, y \in C^s}log(p(\frac{y=o}{x;\theta})),
    \end{split}$}
\end{equation}
where the class probabilities are defined as:
\begin{equation}
    p(\frac{y=s}{x,\theta}) = \frac{\exp(F_v\cdot f_{ts})}{\sum_{(\bar{s})\in A}\exp(F_v \cdot f_{ts})},
\end{equation}
\begin{equation}
    p(\frac{y=o}{x,\theta}) = \frac{\exp(F_v\cdot f_{to})}{\sum_{(\bar{o})\in O}\exp(F_v\cdot f_{to})},
\end{equation}
$f_{ts}$ and $f_{to}$ represent the decomposed attribute and object representations from the text embeddings.
In conclusion, our final loss is computed as the sum of its parts, where $0 < \alpha, \beta, \gamma < 1$
\begin{align}
    \begin{split}
    L =& L_{\text{st+obj}} + \alpha L_{dfm} + \beta \nonumber L_{spm} + \gamma (L_{r} + L_{\text{InfoNCE}}) ~.
    \end{split}
\end{align}
\section{Experiment}
\begin{table*}[!htb]
    \begin{center}
    \fontsize{9}{11}\selectfont
        \begin{tabular}{r|c c c c | c c c c | c c c c}
        \toprule
             \multirow{2}{*}{Method} &  \multicolumn{4}{c|}{MIT-States} & \multicolumn{4}{c|}{UT-Zappos} & \multicolumn{4}{c}{C-GQA}\\
             & S & U &  H & AUC & S & U &  H & AUC & S & U &  H & AUC \\
             \midrule
             CGE \cite{naeem2021learning} & 32.4 & 5.1 & 6.0 & 1.0 & 61.7 & 47.7 & 39.0 & 23.1 & 32.7 & 1.8 & 2.9 & 0.47 \\
             Co-CGEˆClosed \cite{mancini2022learning} & 31.1 & 5.8 & 6.4 & 1.1 & 62.0 & 44.3 & 40.3 & 23.1 & 32.1 & 2.0 & 3.4 & 0.53 \\
             Co-CGEˆOpen \cite{mancini2022learning} & 30.3 & 11.2 & 10.7 & 2.3 & 61.2 & 45.8 & 40.8 & 23.3 & 32.1 & 3.0 & 4.8 & 0.78 \\
             DRANet \cite{li2023distilled}& 29.8 & 7.8 & 7.9 & 1.5 & 65.1 & 54.3 & 44.0 & 28.8 & 31.3 & 3.9 & 6.0 & 1.05 \\
             CLIP \cite{radford2021learning} & 30.1 & 14.3 & 12.8 & 3.0 & 15.6 & 20.5 & 11.3 & 2.2 & 7.5 & 4.4 & 4.0 & 0.28 \\
             CSP \cite{nayak2022learning} & 46.3 & 15.7 & 17.4 & 5.7& 64.1 & 44.1 & 38.9 & 22.7 & 28.7 & 5.2 & 6.9 & 1.2  \\
             DFSP \cite{lu2023decomposed} & 47.5 & 18.5 & 19.3 & 5.8 & 66.8 & 60.0 & 44.0 & 30.3 & \textbf{38.3} & \textbf{7.2} & \textbf{10.4} & \textbf{2.4}\\
             \midrule
            \textbf{HOPE} & \textbf{50.4} & \textbf{19.7} & \textbf{20.7} & \textbf{7.9} & \textbf{68.4} & \textbf{61.9} & \textbf{45.1} & \textbf{31.1} & 35.7 & 6.6 & 9.0 & 2.0\\
            \bottomrule
        \end{tabular}
        \caption{\label{open-world} Open World Evaluation. Comparison to state-of-the-art models}
    \end{center}
\vspace{-.7cm}
\end{table*}
\subsection{Experiment Setup}
\begin{description}[style=sameline, leftmargin=*]
\item[Datasets.]
Our experiments are conducted on three real-world, challenging benchmark datasets: MIT-States, UTZappos, and C-GQA. The MIT-States dataset is frequently used in CZSL research, containing 115 states, 245 objects, and 1962 compositions, and has been embraced by many previous methods. UTZappos includes 50025 images of shoes, featuring 16 states and 12 objects. On the other hand, C-GQA is a newly introduced, large-scale benchmark derived from in-the-wild images, offering more generalized compositions. Being the most popular dataset for CZSL, C-GQA contains 453 states, 870 objects, and a total of 39298 images, which include over 9500 compositions.

\item[Metrics.] 
Following the framework in \cite{mancini2021open}, we evaluate prediction accuracy considering both seen and unseen compositions under both closed-world and open-world scenarios. Specifically, ``Seen" (S) denotes accuracy evaluated solely on known compositions, while ``Unseen" (U) denotes accuracy evaluated exclusively on unknown compositions. Moreover, we calculate the Harmonic Mean (HM) of the S and U metrics. Given the tendency of zero-shot models to favor known compositions, we generate a seen-unseen accuracy curve at various operating points, ranging from a bias of negative infinity to positive infinity, and determine the Area Under the Curve (AUC). 
\item[Implementation Details.]Our HOPE system is implemented using PyTorch version 1.12.1 and is optimized using the Adam optimizer over 10 epochs. The image and text encoders within this model are both constructed on top of the pre-trained CLIP Vit-L/14 architecture. We set the $\alpha, \beta, \gamma$ in the final loss function $L$ to be 0.9, 0.8, and 0.3, respectively. Furthermore, we have configured the Soft MoE with 2 layers. The complete model was trained and evaluated on a single NVIDIA RTX 3090 GPU.
\end{description}

\subsection{Results}
We present experimental comparisons in Table~\ref{closed-world} and Table~\ref{open-world} with previously established compositional zero-shot learning methods, including CGE \cite{naeem2021learning}, Co-CGE \cite{mancini2022learning}, SCEN \cite{li2022siamese}, KG-SP \cite{karthik2022kg}, CSP \cite{nayak2022learning} and DFSP \cite{lu2023decomposed}. The performance is evaluated in both closed-world and open-world scenarios. In the closed-world scenario, our HOPE model sets a new benchmark on the MIT-States and UT-Zappos datasets, displaying superior performance over the recent state-of-the-art method DFSP \cite{lu2023decomposed}. Specifically, on the MIT-States dataset, HOPE surpasses DFSP [20] by increasing the unseen accuracy by 2.6\%, seen accuracy by 3.6\%, harmonic mean by 2.6\%, and AUC by 2.7\%. On the UT-Zappos dataset, HOPE exhibits an improvement of 1.7\%, 2.2\%, 1.9\%, and 1.5\% on seen accuracy, unseen accuracy, harmonic mean, and AUC, respectively, compared to DFSP \cite{lu2023decomposed}. Regarding the C-GQA dataset, the performance of HOPE provides insightful data, prompting a detailed ablation study to better understand the dynamics, as elaborated in the subsequent section.
\section{Ablation Study}
In this section, we conduct extensive experiments on each HOPE's component to identify the key contributing factors to its success.
\subsection{Effectiveness of Memory}
When considering the memory update mechanism during training, its configuration plays a pivotal role in the model's overall effectiveness. Specifically, while our method excels with the MIT-States and UT-Zappos datasets, it struggles with the C-GQA dataset. T-SNE visualizations in Figure~\ref{tsne} of the memory highlight that the C-GQA dataset's memory structure is not as well organized.
\begin{table}[!htb]
    \begin{center}
    \fontsize{9}{11}\selectfont
    \setlength{\tabcolsep}{5pt}
        \begin{tabular}{c |c|c c|c c|c c}
            \toprule
            & Me- & \multicolumn{2}{c|}{1-shot} & \multicolumn{2}{c|}{5-shot} & \multicolumn{2}{c}{full}\\
            & mory & S & U & S & U & S & U\\
            \midrule
            \multirow{3}{*}{MIT-States} & 1 & 32.7 & 45.1 & 40.1 & 49.3 & 50.5 & 54.5 \\
            & 5 & 33.5 & 49.8 & 40.5 & 51.6 & 50.4 & 54.6 \\
            & 10 & \textbf{33.6} & \textbf{51.3} & 40.4 & \textbf{51.7} & 50.5 & 54.6 \\
            \midrule
            \multirow{3}{*}{UT-Zappos} & 1 & 51.3 & 62.0 & 58.8 & 65.5 & 68.4 & 73.6 \\
            & 5 & 54.6 & 65.9 & 60.6 & 67.1 & 68.4 & 73.9 \\
            & 10 & \textbf{55.1} & \textbf{66.7} & \textbf{60.6} & \textbf{67.3} & 68.4 & 73.9 \\
            \midrule
            \multirow{2}{*}{C-GQA} & 1 & n/a & n/a & n/a & n/a & 35.8 & 30.8  \\
            & 5 & n/a & n/a & n/a & n/a & \textbf{35.8} & 
            \textbf{31.3} \\
            \bottomrule
        \end{tabular}
        \caption{\label{memory} Accuracy for Seen (S) and Unseen (U) on MIT-States, UT-Zappos, and C-GQA Datasets. This table details model performance on MIT-States, and UT-Zappos (replicating C-GQA properties) and on C-GQA itself.}
    \end{center}
     \vspace{-3em}
\end{table}

To explore reasons that HOPE underperforms on the C-GQA dataset, we notice that the C-GQA dataset has a significantly low number of images per class, around 2-3 images per class, compared to the C-GQA's, MIT-States and UT-Zappos. We speculate that insufficient data can harm the memory-building process. To prove this, we experimental test the MIT-States and UT-Zappos datasets by limiting the number of images. Table~\ref{memory} illustrates that increasing the number of training data significantly enhances HOPE's accuracy on both seen and unseen metrics. In detail, the growth in seen accuracy is up to 17.8\% and 17.1\% in MIT-States and UT-Zappos datasets, respectively. Meanwhile, unseen accuracy is also elevated up to 9.1\% and 11.9\% on those datasets. 

Additionally, we evaluate the model on different memory settings by changing the number of images for creating general class embeddings. This reveals that improving the memory quality can boost performance. In the 1-shot and 5-shot framework, incrementing the number of images to design the memory gradually enhances HOPE's accuracy. Notably, there is a colossal increase approximately 6.2\% in unseen accuracy by using 10 images instead of a single image for memory initialization. Training with the full dataset converges to the same performance because the memory is differentiable throughout the training process. Therefore, the underperformance in the C-GQA dataset is mainly caused by the fact that many classes are represented by only a few images, limiting the memory's representation fails to capture a comprehensive class essence, which is a mandatory property for the memory's embeddings to be effectively composed.

\subsection{Contrastive Loss Impacts }
The incorporation of contrastive loss presents a distinct training dynamic. Table \ref{contrastive loss ablation} illustrates that without it, the model tends to classify input images using retrieved features as auxiliary support. Introducing contrastive loss shifts the focus towards compositional learning, significantly boosting performance on unseen combinations at the expense of seen ones.  A balanced training strategy that combines both approaches yields the best results. 
\begin{table}[!htb]
    \begin{center}
    \fontsize{9}{11}\selectfont
        \begin{tabular}{r |c c c c}
            \toprule
            \multirow{ 2}{*}{Method} &  \multicolumn{4}{|c}{MIT-States}\\
            & S & U &  H & AUC\\
            \midrule
            without InfoNCE loss& \textbf{47.5} & 52.4 & \textbf{37.8} & \textbf{20.9} \\
            with InfoNCE loss & 45.5 & \textbf{55.6} & 36.5 & 19.8\\
            \bottomrule
        \end{tabular}
        \vspace{0.5em}
        \caption{\label{contrastive loss ablation} Closed world evaluation on different loss configurations}
    \end{center}
    \vspace{-1em}
\end{table}
Figure~\ref{tsne} also shows that contrastive loss also helps maintain feature clusters, preventing the model from updating only those features derived directly from the CLIP loss.
\subsection{Modern Hopfield Network Robustness}
To assess the robustness of MHN in pattern retrieval, Table~\ref{mhn} shows the accuracy of MHN in retrieving seen compositions from the test dataset and how the retrieval loss influences the accuracy. For novel (unseen) compositions, we track the number of successful retrievals featuring specific attributes and objects from memories $M_v^a$ and $M_v^o$, respectively. \cite{li2024context} shows that the unseen composition of a state and an object can be easily adapted from their synonym. For eg, a red truck can be inferred from a red car, so we likewise evaluate MHN's ability to retrieve synonyms of attributes or objects using the NLTK library.
\begin{table}[!htb]
    \begin{center}
    \setlength{\tabcolsep}{5pt}
    \fontsize{9}{11}\selectfont
        \begin{tabular}{r|c c c}
            \toprule
            Retrieval & \multicolumn{3}{c}{MIT-States}\\
            Loss & Seen & Unseen & Unseen + Synonyms \\
            \midrule
            With & 91.6\% & 69.3\%  & 78.2\% \\ 
            Without & 91.3\%& 54.9\% & 65.1\%  \\ 
            \bottomrule
        \end{tabular}
        \caption{\label{mhn} Performance Analysis of MHN: The retrieval accuracy of MHN is assessed on the MIT-States dataset under various conditions. Accuracy is shown for compositions seen during training, unseen compositions, and unseen compositions inclusive of synonyms, with and without the application of retrieval loss.}
    \end{center}
    \vspace{-2em}
\end{table} 

The retrieved seen accuracy is stable at 91.3\% regardless of the retrieval loss, demonstrating the outstanding inherent retrieval capability of the Modern Hopfield Network. Furthermore, in unseen scenarios where the model can only search for similar attributes and objects, there is a marginal decline from 69.3\% down to 54.9\% when removing the retrieval loss. Moreover, we found that the Modern Hopfield Network retrieves the attributes and objects that are synonyms to the ground-truth label. By including the synonyms in evaluation metrics, the accuracy gained by 8.9\%, emphasizing the robustness of this component in retrieving similar composition embeddings. 
\subsection{Analysis of Expert Allocation in Soft MoE}
\vspace{-.2cm}
\begin{figure}[!htb]
    \centering
    \includegraphics[width=\columnwidth,keepaspectratio]{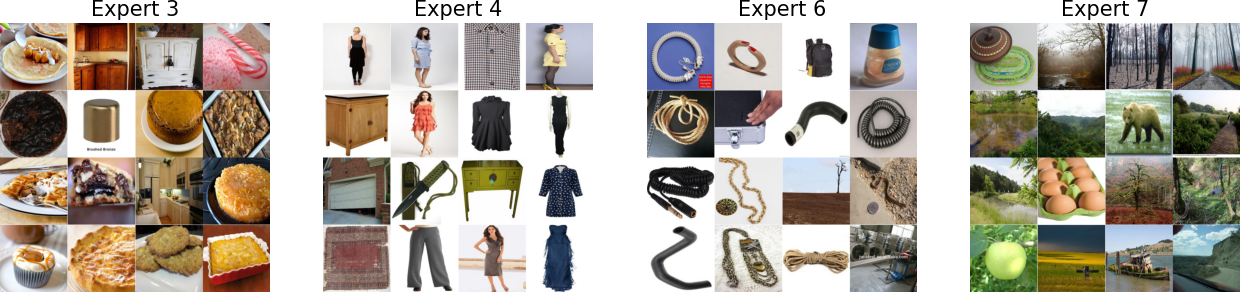}
    \caption{Visualization of Soft Mixture of Experts. Each `expert' focuses on distinct categories of data – Expert 3 on food items, Expert 4 on apparel, etc.– demonstrating the model's ability to assign and weigh inputs across different neural network sub-models for enhanced specialization and accuracy in classification tasks}
    \label{moe-visualize}
\vspace{-1em}
\end{figure}
Our analysis examines how the Soft Mixture of Experts (Soft MoE) distributes inputs based on their attributes or objects. By contrasting this with a cross-attention approach, table \ref{moe_rebut} shows that implementing Soft MoE enhances accuracy in all metrics. Visual inspection in Fig~\ref{moe-visualize} reveals that the Soft MoE tends to assign images with similar attributes or objects to the same experts, indicating a degree of efficient routing. Yet, this pattern is confined to a few experts, with many experts still handling a mix of unrelated inputs. This selective engagement of experts is noteworthy, as it implies that the model does not engage all patterns or features equally. Instead, it adopts a tailored strategy for image recognition, focusing on distinctive features to potentially improve its performance on new, unseen data.
\begin{table}[!htb]
    \begin{center}
        \fontsize{9}{11}\selectfont
        \begin{tabular}{r|c c c c}
            \toprule
            \multirow{ 2}{*}{Methods} &  \multicolumn{4}{c}{MIT-States}\\
            & S & U &  H & AUC\\
            \midrule
            Self Attention\cite{vaswani2017attention} & 48.9 & 53.1 & 38.4 & 22.2 \\
            Soft MoE\cite{puigcerver2023sparse} & \textbf{50.5} & \textbf{54.6} & \textbf{39.9} & \textbf{23.3} \\ 
            \bottomrule
        \end{tabular}
        \caption{\label{moe_rebut} Comparative closed-world evaluation results of Self Attention and Soft Mixture of Experts methods on the MIT-States dataset,  after 10 training epochs.} 
    \end{center}
    \vspace{-2.5em}
\end{table}
\section{Conclusion}
In this paper, we propose HOPE, a novel framework for Compositional Zero-Shot Learning (CZSL) that mimics human adaptability to new state-object combinations. By integrating the Modern Hopfield Network with a Mixture of Experts, our framework effectively recalls and assembles relevant primitives into novel compositions. Additional loss functions are proposed to ensure that each part of the system contributes optimally to information recall and association. Our extensive evaluations and detailed analyses establish the superior performance of HOPE across various standard datasets compared to SOTA and provide insights into its limitations with the C-GQA dataset.

\noindent\textbf{Acknowledgements}
The authors appreciate the support of the ``Opportunity for Undergraduate Research" funding and Ms. Dinh Mai Phuong from VinUniversity for providing essential facilities to conduct this work.

{\small
\bibliographystyle{ieee_fullname}
\bibliography{main}
}

\end{document}